\documentclass[10pt,conference]{IEEEtran}
\IEEEoverridecommandlockouts
\usepackage{cite}
\usepackage{amsmath,amssymb,amsfonts}
\usepackage{algorithmic}
\usepackage{graphicx}
\usepackage{textcomp}
\usepackage{xcolor}
\def\BibTeX{{\rm B\kern-.05em{\sc i\kern-.025em b}\kern-.08em
    T\kern-.1667em\lower.7ex\hbox{E}\kern-.125emX}}

\begin{document}
\bibliographystyle{IEEEtran}

\title{An Overview and Discussion on Using \\ Large Language Models for Implementation Generation of Solutions to Open-Ended Problems 
}

\author{\IEEEauthorblockN{Hashmath Shaik}
\IEEEauthorblockA{\textit{Department of ECE} \\
\textit{Stony Brook University}\\
Stony Brook, NY 11794-2350 \\
hashmath.shaik@stonybrook.edu}
\and
\IEEEauthorblockN{Alex Doboli}
\IEEEauthorblockA{\textit{Department of ECE} \\
\textit{Stony Brook University}\\
Stony Brook, NY 11794-2350 \\
alex.doboli@stonybrook.edu}
}
\maketitle

\begin{abstract}
Large Language Models offer new opportunities to devise automated implementation generation methods that can tackle problem solving activities beyond traditional methods, which require algorithmic specifications and can use only static domain knowledge, like performance metrics and libraries of basic building blocks. Large Language Models could support creating new methods to support problem solving activities for open-ended problems, like problem framing, exploring possible solving approaches, feature elaboration and combination, more advanced implementation assessment, and handling unexpected situations. This report summarized the current work on Large Language Models, including model prompting, Reinforcement Learning, and Retrieval-Augmented Generation. Future research requirements were also discussed. 
\end{abstract}


\section{Introduction}

Problem solving is the process of creating a solution for a problem description~\cite{Fiore2010, fischer2012process, Sun2020, Wiltshire2018}. The solution can be an {\em explanation} for a set of properties exhibited by a static or dynamic situation, e.g., a mathematical proof, or an {\em implementation} (realization), which is the construction of a new materialization (e.g., design) that exhibits the required properties as a result of their operation (functioning, execution). This report focuses on the implementation (realization) of problem solving.

Creating an implementation can pertain to the three general-purpose problem-solving situations: well-defined problems, ill-defined problems, and open-ended problems~\cite{schraw1995cognitive, Doboli2014, Doboli2015}:
\begin{enumerate}
\item
Well-defined problem solving for implementation construction describes situations in which an existing solution can be reused with some incremental changes to solve a new problem. For example, textbook algorithms are utilized to solve a new problem by selecting proper data structures and customizing the algorithm parameters, like the conditions of conditional statements and the iterations of loops. Using parameterized templates for circuit design~\cite{koza1997reuse, Wirfs2006, Tang2006, Wei2007} belongs to this category too.         

\item 
Ill-defined problem solving for implementation construction represents cases in which the existing implementations cannot solve all requirements, i.e. they satisfy some but not others~\cite{klein1978improvement, leighton1999assessment}. Changing the parameters of the implementation does not address the issue. Problem solving includes options, like producing a description of the implementation trade-offs by parameter sampling and selecting the best compromise, exploring implementation alternatives for specific fragments of the implementation, so that better trade-offs result for the overall solution, and selecting a different approach (principle) for an implementation, including situations when a new implementation must be built, similar to open-ended solving for building a new implementation. 

\item
Open-ended problem solving for implementation generation requires devising new solutions with a significant departure and characteristics from previous implementations. The understanding of this process is still limited~\cite{doboli2021novel, wang2020enhanced}. Also, there are insufficient metrics to describe the degree to which the process is systematically progressing towards success, e.g., building a new implementation. Typical activities include problem framing and problem understanding, identifying and selecting the solving approach, divide and conquer (e.g., problem partitioning into sub-problems), implementation elaboration through trial-end-error, feature combination, adjustment, abstraction and insight gaining, implementation analysis to find pros and cons and the impact of features on the implementation operation, implementation modification, error correction, and handling unexpected situations. 
\end{enumerate}

As summarized in the next section, traditional automated implementation generation focuses mainly on elaboration and parameter trade-off exploration, for which the domain knowledge of the implementation is captured by customized metrics~\cite{Aho2006 } or in a library of basic building blocks~\cite{Aho2006, Doboli2004}. The library is static and does not evolve to incorporate new knowledge either from external sources or as a byproduct of implementation generation. Moreover, traditional methods assume the existence of a problem specification expressing at least functional and  performance requirements, but more often the algorithm or architecture (structure) of the implementation~\cite{Fingeroff2010, Doboli2004}. Hence, it can be argued that existing methods focus mainly on well-defined and ill-defined problems but less on implementation generation for open-ended problem solving. Existing approaches cannot tackle problem framing and exploring solution approaches, even though trial-and-error and rapid prototyping are essential in understanding new opportunities and limitations. Moreover, there is little automated support for divide and conquer and architecture creation, combination of features from different solutions, and handling unexpected situations. In general, traditional methods struggle with any activity conducted at a level above an algorithmic description of an implementation. 

However, recent advances in Large Language Models (LLMs) created opportunities to devise novel automated implementation generation methods that can tackle problems beyond algorithmic specifications and may use domain knowledge that is dynamically learned over time. Arguably, LLMs could contain knowledge that is continuously updated by learning new features either from external documents or based on their own previously generated implementations. Implementation assessment could be improved by comparing it to similar, externally available implementations and considering collective feedback and preferences expressed for other solutions. The opportunities and limitations of an implementation can be better understood by embedding it into the trend of related designs. Moreover, support can be offered for problem framing and exploring possible solution approaches, activities that are often collective, in a team. LLMs can process multi-modal descriptions, including natural language and images with certain degrees of specification completeness, unknowns, and ambiguity. Hence, understanding the capabilities of LLMs for implementation generation, possibly in conjunction with traditional methods, is required. These capabilities mostly emerge from LLMs being able to learn a broad range of associations in multi-modal data and diverse contexts.    

This report studied the degree to which LLMs, possibly using prompting, Reinforcement Learning (RL) and Retrieval-Augmented Generation (RAG), can model the activities of implementation generation for an open-ended problem solving. The goal was to identify how LLMs and their extensions can contribute to implementing problem-solving activities that are not addressed in traditional methods. The report offers an extensive presentation of prompting methods, RAG techniques, and RL approaches. Then, the using of LLMs to implement problem-solving activities not available in traditional automated implementation generation was discussed. New research requirements were also offered. The report argued that these requirements refer to topics, like constructing the implementation approach, effectively controlling elaboration, robust qualitative and quantitative assessment across abstraction levels, knowledge memorizing during learning, and managing the problem solving process. 

The report has the following structure. Section~II offers an overview of the work on traditional, automated implementation generation. Section~III presents an overview of LLMs. Section~IV discusses the similarities of LLMs and traditional automated implementation generation methods and summarizes the related research needs. Conclusions end the report.

\section {Overview of Traditional Automated Implementation Generation}

Traditional approaches to automatically generate implementations can be grouped into four broad categories: (i)~approaches based on high-level specifications, (ii)~methods using evolutionary algorithms, (iii)~agent-based methods, and (iv)~cognitive architectures. The four categories are summarized next. 

{\bf (i)~Approaches based on high-level specifications:} These approaches include traditional compiling methods to generate executable code~\cite{Aho2006} and high-level synthesis methods~\cite{Fingeroff2010, McConaghy2009, Doboli2003a, Doboli2003b} and template-based synthesis~\cite{Tang2006, Wei2007} to create electronic circuits and systems. They use high-level specifications described using a programming language. Conceptually, specifications serve as parameterized descriptions of the target implementation architecture. Specifically, internal representations are built using a set of predefined rules (e.g., language grammar) applied to the specifications and then used to create an optimized hardware design by exploring different optimization possibilities. Prediction models or simulation tools are integrated to evaluate the performance of possible implementation alternatives. 

These methods address the problem-solving activities in the following ways: The specification gives an unambiguous, complete description of the parameterized architecture. Thus, there is no problem framing step and problem understanding is fully addressed during specification creation. Divide and conquer is defined by the structuring of the specification. Also, there is no step of exploring possible implementation alternatives, as the specification explicitly describes the data processing steps, including the connections between the sequences of processing steps, i.e. using the processing outputs as inputs for the next processing steps. Hence, feature combination during elaboration only connects predefined operators which do not change their function based on the connections. From the point of view of cognitive psychology, these combinations are relation-based combinations but do not reflect feature-based combinations, in which features of a concept are transferred to another concept~\cite{Wisniewski1997}. Hence, there are no unexpected situations, including emerging features. Implementation analysis uses performance models and simulation, even though the pros and cons of an implementation are rarely causally linked to the implementation fragments responsible for them. Hence, the insight gain is limited. Trial-and-error (possibly guided by priority functions), implementation modification, and adjustment are only at the level of optimizing the architecture parameters. There is no abstraction or summarization during the process. Error correction requires to modify the specification and then repeat the problem-solving process. 

{\bf (ii)~Methods using evolutionary algorithms:} These methods create a dynamic process, in which large populations of solutions originate new populations through traditional operators, i.e. selection, crossover, and mutation~\cite{Kruiskamp1995}. Selection means propagating high-fitness individuals from the current to the next population, crossover combines features of a set of solutions to produce new solutions, and mutation randomly changes solution features. 

These methods do not include problem framing and understanding. Identifying and selecting the implementation approach has been studied less, even though it is possible to maintain separate sub-populations, each for a different approach, and then giving higher priority to the sub-populations that include more high-quality implementations. There is no divide-and-conquer to separate a problem into sub-problems and no explicit error correction. Trial-and-error is mimicked through the mutation operator, even though mutation does not implement a systematic exploration process guided by the learned knowledge. There is no insight gaining during the process, abstraction or summarization of the learned knowledge, and no explicit identification of unexpected situations. Crossover implements combination, including feature and relation combination. Similar to the previous category, implementation analysis uses performance models and simulation to produce a fitness value that controls the selection of the better implementations. However, there is no explicit identification of the causal features that produce the pros and cons of an implementation, thus there is no implementation adjustment, modification, or correction guided by causal information. There is no explicit memory mechanism, features being implicitly memorized through a population, and there is no possibility to backtrack to previous states to attempt exploring a different path.  

{\bf (iii)~Agent-based methods}: These methods utilize multiple interacting agents, each agent having its own memory and running its own decision-making algorithm~\cite{Chopra2013, Bonabeau2002}. Even though traditional agents realize simple decision-making algorithms, e.g., through a set of simple rules in response to specific inputs, it is possible to consider more complex methods, such as each agent running its own synthesis algorithm or population-based evolution. Agents interact with each other by communicating high-quality implementations and features, or implementation steps, which then can be utilized by the other agents, too. 

Depending on their decision-making procedure, agent-based methods have similar characteristics, like the methods of the previous two categories. Their main advantage is their capacity to simultaneously maintain multiple perspectives about the implementation creation process, e.g., through their local memory, preferences, priorities, etc., and then aggregate these perspectives to improve problem solving. It can be argued that they mimic the implementation creation process by a team (team problem solving)~\cite{Lapp2017, doboli2021novel}.   

{\bf (iv)~Cognitive architectures}: Cognitive architectures (CAs) mimic the brain activities during problem solving~\cite{Anderson1996, Laird2012, Rosenbloom2016, Kieras1997, Sun2003}. Architectures include modules for knowledge representation, knowledge memory, knowledge classification, summarization, comparison, decision-making, prediction, learning, and goal setting. For example, SOAR CA models cognition-based problem solving~\cite{Laird2012}, using operation selection and application (e.g., state elaboration, operator proposal and evaluation, and decision). Knowledge is procedural if-then rules selected through matching. Learning stores short-cuts to solutions, conditions for applying the rules, and utility updates. ACT-R CA uses multiple symbolic knowledge representations, declarative and procedural information learning, and utility-based decision making~\cite{Anderson1996}. EPIC CA matches in parallel production rules to the working memory, followed by the selection of firing rules for multiple goals~\cite{Kieras1997}. Sigma CA includes mixed symbolic-probabilistic, discrete-continuous representations, knowledge summarization and integration, and inference-based reasoning~\cite{Rosenbloom2016}. Clarion CA maintains explicit and implicit cognition, each having different representations and processing methods, e.g., rule extraction, generalization, specialization, backpropagation, and reinforcement learning~\cite{Sun2003}. InnovA is a CA for automated design of electronic circuits~\cite{Li2018}.

\section {Overview of Large Language Models, Prompt Engineering, Retrieval-Augmented Generation, and Reinforcement Learning}

\subsection {Large Language Models}

Large Language Models (LLMs), primarily those built on transformer architectures, have made significant strides in producing coherent, contextually relevant text~\cite{vaswani2017attention}. They excel at pattern recognition and can generate fluent natural language by leveraging billions of parameters trained on massive corpora~\cite{brown2020language}. However, their computational principle—self-attention over sequential data—imposes fundamental limitations that hinder their ability to perform the rich, open-ended problem-solving tasks described in the previous sections.

At the core of these limitations is the reliance on statistical correlations rather than genuine logical or conceptual understanding. While self-attention excels at identifying relevant tokens in a sequence, it does not inherently encode hierarchical structures, domain-specific causal rules, or strict logical constraints. This stands in contrast to open-ended problem solving, where the concept space can be segmented into three main categories—hierarchical concepts, alternative concepts, and fundamental concepts—and the action space encompasses complex operations, such as feature combination, dynamic adjustment, abstraction, insight generation, and summarization ~\cite{bengio2013representation}. LLMs struggle to engage these conceptual spaces in a principled way because they are not grounded in mechanisms that ensure hierarchical reasoning, strategic problem decomposition, or the flexible reuse of insights and intermediate representations~\cite{lake2017building}.

Another critical shortcoming is that LLMs tend to produce generalized answers aligned with the statistical patterns seen in their training data ~\cite{weidinger2021ethical}. They are not inherently equipped to execute a true divide-and-conquer approach to complex tasks, nor can they systematically apply trial-and-error strategies. For example, while open-ended problem solving may demand iterative refinement—where a solver explores a space of possible solutions, backtracks as necessary, and learns from failed attempts—an LLM’s output is typically a single forward pass~\cite{bubeck2023sparks}. Without an internal model of logical inference, memory structures that accumulate knowledge over multiple steps, or explicit strategy formulations, LLMs cannot easily correct their reasoning or adapt their approach based on previous mistakes~\cite{zellers2019defending}. This leads to issues such as hallucinations, where models confidently assert falsehoods; distractions, where irrelevant details are emphasized; and a general inability to build complex, causally grounded explanations.

Some researchers have explored techniques like constraint-based decoding to enforce logical or linguistic rules at inference time~\cite{post2018fast}. This can improve consistency and coherence to some extent, but it remains an add-on rather than a fundamental solution. Constraint-based methods do not grant the model a deeper conceptual understanding; they merely prune outputs that violate predetermined constraints. Similarly, improvements like sparse attention mechanisms reduce computational complexity, adapter layers can inject domain-specific knowledge ~\cite{houlsby2019parameter}, and memory-augmented transformers attempt to store and reuse intermediate reasoning steps. While these approaches enhance performance on certain tasks, they do not fully overcome the inherent limitations of attention-based architectures or enable robust open-ended problem solving. The models are still limited by their training data, biased toward patterns present therein, and lack the ability to intentionally search concept space, systematically test hypotheses, or derive new conceptual abstractions beyond what is statistically suggested~\cite{geirhos2020shortcut}.

In response to these challenges, a body of methods has emerged to push LLMs closer toward more sophisticated reasoning and problem-solving behaviors. This work can be broadly divided into three interrelated categories: Prompt Engineering, knowledge retrieval through Retrieval-Augmented Generation, and model refinement through Reinforcement Learning).

\subsection{Prompt Engineering}

Prompting techniques utilize carefully constructed input prompts to guide the model’s response generation process. Techniques can be grouped into five categories dicussed next. 


{\em a) Single-stage prompting (SSP)}: SSP methods directly instruct the model without iterative refinement. 
Meanwhile, Basic + Annotation Guideline-Based Prompting + Error Analysis-Based Prompting~\cite{hu2024improving} uses formally defined entity annotation guidelines to specify how clinical terms should be identified and categorized, ensuring clarity in entity recognition. In addition, it incorporates instructions derived from analyzing common model errors, such as addressing ambiguous entity boundaries or redefining prompts for overlapping terms. This strategy significantly improves clinical Named Entity Recognition, with relaxed F1 scores reported as 0.794 for GPT-3.5 and 0.861 for GPT-4 on the MTSamples dataset ~\cite{mtsamples} and 0.676 for GPT-3.5 and 0.736 for GPT-4 on the VAERS dataset ~\cite{vaers}, demonstrating its effectiveness.

{\em b) Reasoning strategies:} These methods are of three types: linear, branching, and iterative reasoning.

Linear reasoning methods such as Chain-of-Thought (CoT), Complex CoT, Thread-of-Thought (ThoT), Chain-of-Knowledge (CoK), Chain-of-Code (CoC), Logical Thoughts (LoT), Chain-of-Event (CoE), and Chain-of-Table generate a single, step-by-step sequence (chain) of responses toward the final answer. Methods differ in the type of task they target, i.e., code generation, summarization, and logical inference, and in how they refine or represent intermediate steps. CoT shows that using intermediate prompting steps can enhance accuracy, e.g., up to 39\% gains in mathematical problem solving~\cite{wei2022chain}. An example of in-context prompt for CoT might be: \textit{“If the problem is ‘Calculate 123 × 456,’ break it down as (100 + 20 + 3) × 456 and compute step-by-step.”} Complex CoT uses more involved in-context examples, improving performance by as much as 18\% on harder tasks~\cite{fu2022complexity}. ThoT tackles long or chaotic contexts by \textit{breaking them into manageable parts} (e.g., dividing long passages into sections for sequential summarization)~\cite{zhou2023thread}, while CoK strategically adapts and consolidates knowledge from multiple sources to ensure coherence and reduce hallucination~\cite{li2023chain}. CoC specializes in code-oriented reasoning by simulating key code outputs (e.g., predicting intermediate variable states for debugging)~\cite{li2023chain1}, whereas LoT integrates logical equivalences and \textit{reductio ad absurdum} checks to refine reasoning chains (e.g., validating statements by identifying contradictions in their negations)~\cite{zhao2023enhancing}. CoE handles summarization by extracting, generalizing, filtering, and integrating key events (e.g., pinpointing main events from news articles)~\cite{bao2024chain}, and Chain-of-Table extends CoT techniques to tabular data, dynamically applying transformations like filtering or aggregation to generate coherent answers~\cite{wang2024chain}.

Branching reasoning methods, like Self-Consistency, Contrastive CoT (or Contrastive Self-Consistency), Federated Same/Different Parameter Self-Consistency/CoT (Fed-SP/DP-SC/COT), Tree-of-Thoughts, and Maieutic Prompting, explore multiple possible reasoning paths in parallel. Branching techniques vary in how they sample or fuse paths, some relying on consensus votes and others on dynamic adaptation or tree-based elimination.
Self-Consistency, for instance, samples diverse solution paths and selects the most consistent final answer, achieving gains of over 11\% on math tasks~\cite{wang2022self}. Contrastive CoT incorporates both correct and incorrect in-context examples to broaden the model’s understanding, improving performance by over 10\% compared to standard CoT~\cite{chia2023contrastive}. Fed-SP-SC leverages paraphrased queries to crowdsource additional hints~\cite{liu2023federated}, while ToT maintains a tree of partial solutions and systematically explores them with breadth-first or depth-first strategies, offering up to 65\% higher success rates than CoT on challenging math tasks(ToT)~\cite{yao2024tree}. Maieutic Prompting likewise generates a tree of propositions to reconcile contradictory statements, surpassing linear methods by 20\% on common-sense benchmarks~\cite{jung2022maieutic}.


Iterative reasoning approaches, such as Plan-and-Solve (PS), Program-of-Thoughts (PoT), Chain-of-Symbol (CoS), Structured Chain-of-Thought (SCoT), and Three-Hop Reasoning (THOR), refine solutions step by step, often by passing intermediate outputs back into the model to enhance accuracy. PS explicitly decomposes tasks into planning and execution phases, where the planning phase structures the problem into smaller sub-tasks, and the execution phase solves them sequentially. This reduces semantic and calculation errors, outperforming Chain-of-Thought (CoT) prompting by up to 5\%~\cite{wang2023plan}. PoT enhances performance by separating reasoning from computation: the model generates programmatic solutions executed by a Python interpreter, achieving up to 12\% accuracy gains in numerical and QA tasks~\cite{chen2022program}. CoS encodes spatial and symbolic relationships using concise symbolic representations, which improves reasoning in spatial tasks by up to 60.8\%~\cite{hu2023chain}. SCoT introduces structured reasoning through program-like branching and looping, significantly improving code generation accuracy by up to 13.79\%~\cite{li2023structured}. Finally, THOR tackles emotion and sentiment analysis through a three-stage approach: aspect identification, opinion analysis, and polarity inference. This structured method achieves superior performance compared to previous supervised and zero-shot models~\cite{fei2023reasoning}. These approaches exemplify the power of iterative methods in breaking complex problems into manageable components, thereby reducing errors and improving overall performance.


{\em c) Multi-Stage Prompting (MSP)}: MSP techniques rely on iterative feedback loops or ensemble strategies. MSP methods systematically refine outputs and incorporate multiple response paths, e.g., through voting or iterative analysis, to yield more robust and accurate solutions, particularly in domains requiring deeper reasoning or tailored task adaptation. Ensemble Refinement (ER) ~\cite{singhal2023towards} builds on Chain-of-Thought (CoT) and Self-Consistency by generating multiple CoT-based responses at high temperature (introducing diversity) and then iteratively conditioning on generated responses to produce a more coherent and accurate output, leveraging insights from the strengths and weaknesses of initial explanations and majority voting. Auto-CoT~\cite{zhang2022automatic} constructs demonstrations automatically by clustering queries from a dataset and generating reasoning chains for representative queries using Zero-Shot-CoT. Clustering is achieved by partitioning questions into groups based on semantic similarity, ensuring that representative queries capture the diversity of the dataset. ReAct~\cite{yao2022react} interleaves reasoning traces—thought processes that explain intermediate steps—with action steps that execute operations, enabling superior performance in complex tasks by seamlessly combining reasoning and action. Moreover, Active-Prompt~\cite{diao2023active} adaptively selects the most uncertain training queries, identified via confidence metrics like entropy or variance, for human annotation, boosting few-shot learning performance by focusing on areas with the highest uncertainty.

{\em d) Knowledge Enhancement}: These approaches use high-quality examples and strategic self-monitoring to improve LLM performance. They pertain to two types, example-based and meta-level guidance methods

Example-based methods leverage auxiliary examples or synthesized instances to guide the response creation process of LLMs. MathPrompter~\cite{imani2023mathprompter} focuses on creating a symbolic template of the given mathematical query, solving it analytically or via Python, and then validating the derived solution with random variable substitutions before finalizing the answer. The approach boosts accuracy from 78.7\% to 92.5\%. Analogical Reasoning~\cite{yasunaga2023large} prompts LLMs to generate and solve similar examples before addressing the main problem, resulting in a 4\% average accuracy gain across various tasks. Synthetic Prompting~\cite{shao2023synthetic} involves a backward step, where a new query is generated from a self-constructed reasoning chain, and a forward step, where this query is re-solved; this strategy selects the most complex examples for few-shot prompts, leading to up to 15.6\% absolute improvements in mathematical problem solving, common-sense reasoning, and logical reasoning.


Meta-Level Guidance (MLG) methods enhance LLMs by promoting self-reflection and focusing on pertinent information, thereby reducing errors. Self-Reflection involves the model evaluating its own outputs to identify and correct mistakes, leading to improved performance. For example, in translation tasks, self-reflection enables LLMs to retrieve bilingual knowledge, facilitating the generation of higher-quality translations. Focusing is achieved through techniques like System 2 Attention (S2A) ~\cite{weston2023system}, which filters out irrelevant content by prompting the model to regenerate the context to include only essential information before producing a final response. This two-step approach enhances reasoning by concentrating on relevant details, thereby improving accuracy. S2A has been shown to outperform basic prompting methods, including Chain-of-Thought (CoT) and instructed prompting, particularly on truthfulness-oriented datasets. Metacognitive Prompting (MP) ~\cite{wang2023metacognitive} introduces a five-stage process to further enhance LLM performance: (1)~Comprehension: The model attempts to understands the input, ensuring clarity before proceeding; (2)~Preliminary Judgment: An initial assessment is made based on the understood information; (3)~Critical Evaluation: The initial judgment is scrutinized, considering alternative perspectives and potential errors; (4)~Final Decision with Explanation: A conclusive decision is reached, accompanied by a rationale to support it; and (5)~Self-Assessment of Confidence: The model evaluates its confidence in the final decision, reflecting on the reasoning process. This structured approach enables LLMs to perform consistently better than methods like CoT and Program Synthesis (PS) across various natural language processing tasks, including paraphrasing, natural language inference, and named entity recognition.

{\em e) Task Decomposition}: These approaches break down complex tasks into smaller steps but vary in how they orchestrate and execute the sub-problems. They include problem breakdown and sequential solving methods.


Problem Breakdown approaches include the Least-to-Most method~\cite{zhou2022least}, which addresses the challenge of Chain-of-Thought (CoT) failing on problems more difficult than its exemplars by first prompting the LLM to decompose a query into sub-problems and then solving them sequentially, demonstrating notable improvements over CoT and basic prompting on tasks like commonsense reasoning and mathematical problem solving. The decompositions are characterized by their hierarchical structure, breaking down complex problems into simpler, manageable sub-tasks that build upon each other to facilitate step-by-step reasoning. Decomposed Prompting (DecomP) breaks complex tasks into simpler sub-tasks, each handled with tailored prompts or external tools, ensuring efficient and accurate execution. For instance, the task "Concatenate the first letters of words in 'Jack Ryan'" is decomposed into extracting words, finding their first letters, and concatenating them~\cite{khot2022decomposed}. DecomP leverages modular decomposers to partition problems hierarchically or recursively, assigning sub-tasks to specialized LLMs or APIs. This approach achieves a 25\% improvement over CoT and Least-to-Most methods in Commonsense Reasoning. Program-Aided Language Models (PAL)~\cite{gao2023pal} further leverage interleaved natural language and programmatic steps to enable Python-based execution of the reasoning process, surpassing CoT and basic methods for mathematical and commonsense tasks.



Sequential Solving includes methods, like Binder and Dater algorithms. Binder~\cite{cheng2022binding} integrates neural and symbolic parts by using an LLM both as a parser and executor for natural language queries, leveraging programming languages like Python or SQL for structured execution. Binding is achieved through a unified API that enables the LLM to generate, interpret, and execute code using a few in-context examples, leading to higher accuracy on table-based tasks compared to fine-tuned approaches. Dater~\cite{ye2023large} focuses on few-shot table reasoning by splitting a large table into relevant sub-tables, translating complex queries into SQL sub-queries, and combining partial outcomes into a final solution. These three steps aim to systematically extract meaningful data, execute precise operations, and integrate results to address complex queries, outperforming fine-tuned methods by at least 2\% on Table-Based Truthfulness and 1\% on Table-Based QA, and surpassing Binder on these tasks.

\subsection{Retrieval-Augmented Generation}

Retrieval-Augmented Generation (RAG) addresses one of the major issues of LLMs, which are their lack of a persistent, reliable memory and factual grounding~\cite{lewis2020retrieval}. RAG methods integrate external knowledge sources into the generation process. Instead of relying solely on learned representations within the model’s parameters, the system retrieves relevant documents, facts, or structured data at inference time and incorporates this information into its output. This grounding reduces hallucinations, ensures that the model’s reasoning steps reference accurate and up-to-date information, and can improve the alignment of the solution with real-world constraints~\cite{bechard2024reducing}. The versatility of RAG has led to significant advancements in various domains, such as healthcare, finance, education, and scientific research facilitated by novel frameworks tailored to address challenges in reasoning, problem-solving, and knowledge integration. This review categorized these advancements into four areas: task-specific and schema-based techniques, self-aware and adaptive mechanisms, long-term memory integration, and multi-hop and multi-modal reasoning. The four areas are discussed next.


{\em a) Task-Specific and Schema-Based Retrieval (TSR)}: TSR approaches leverage structured methods to solve problems in domains such as mathematics and knowledge-intensive tasks. Schema-Based Instruction Retrieval-Augmented Generation (SBI-RAG)~\cite{dixit2024sbi} employs schema-based instruction to solve math word problems by predicting relevant schemas, offering a structured problem-solving paradigm. For instance, given the problem, “If a worker earns \$20 per hour, how much will they earn in 10 hours?”, the model predicts the multiplicative schema, which involves calculating the product of hourly earnings and hours worked. Using this schema, the problem-solving process is guided step-by-step: the model multiplies the hourly rate (\$20) by the number of hours (10), resulting in a total earning of \$200. Schemas act as templates for organizing and applying domain-specific knowledge and are inherently tied to knowledge graphs that map relationships between concepts, further enhancing reasoning capabilities. By aligning the problem context with predefined patterns, SBI-RAG ensures systematic and accurate solutions while improving explainability. Similarly, Knowledge Graph-Enhanced RAG Framework (KRAGEN)~\cite{matsumoto2024kragen} employs advanced prompting techniques, notably the graph-of-thoughts (GoT) method, to dynamically decompose complex problems into smaller subproblems. Each subproblem is addressed using relevant knowledge retrieved through the RAG framework, minimizing hallucinations and enhancing solution accuracy. The individual solutions are then consolidated to form a comprehensive answer, with KRAGEN's graph visualization enabling users to interact with and assess the quality of the solution's GoT structure and logic~\cite{matsumoto2024kragen}. These techniques stand out for their ability to address domain-specific challenges while ensuring adaptability through schema-guided reasoning. The use of schemas not only structures the solution process but also facilitates explainability.

In data-driven tasks, Generative Retrieval-Augmented Matching (GRAM) addresses schema matching by employing a hierarchical classification model that dynamically generates prompts for matching attributes across schemas. Specifically, GRAM utilizes a two-step process: first, it performs a coarse-grained classification to identify potential attribute matches. For instance, given two schemas, GRAM might preliminarily match the attribute “Customer Name” in one schema with “Client Name” in another. Then, it refines these matches through fine-grained classification, analyzing the context and patterns in the data to confirm the match and enhance the precision of schema alignment. In this case, GRAM would validate that “Customer Name” and “Client Name” indeed refer to the same entity by assessing their usage and data properties. The prompt generation process, guided by LLMs, enables zero-shot and few-shot learning, allowing GRAM to perform efficiently and accurately in database integration tasks, even when minimal labeled data is available~\cite{liu2024gram}. Similarly, TableRAG~\cite{chen2024tablerag} focuses on reasoning over tabular data by retrieving and processing row-column relationships to interpret structured datasets accurately. It conducts reasoning by leveraging query expansion combined with schema and cell retrieval to pinpoint crucial information before providing it to the language models, enabling efficient data encoding and precise retrieval. This approach allows TableRAG to handle large-scale tables effectively, reducing prompt lengths and mitigating information loss during the reasoning process~\cite{chen2024tablerag}.

{\em b) Self-Aware and Adaptive Retrieval:} Recent RAG frameworks emphasize self-awareness and adaptive mechanisms to address uncertainties in LLMs. Self-aware Knowledge Retrieval (SeaKR)~\cite{yao2024seakr} activates retrieval during high uncertainty and re-ranks snippets to ensure reliability. Specifically, SeaKR~\cite{yao2024seakr} addresses uncertainties arising from the LLM's internal state inconsistencies, triggering retrieval when the model's self-assessed confidence is low. The re-ranking process involves selecting knowledge snippets that most effectively reduce the model's uncertainty, thereby enhancing response accuracy~\cite{yao2024seakr}. Self-RAG~\cite{asai2023self} introduces iterative refinement, where retrieval queries generated during the response process enable reassessment and improvement of outputs. This reassessment involves evaluating the relevance of retrieved information during generation, allowing the model to iteratively refine its responses for enhanced accuracy. Critic-Guided Planning (CR-Planner)~\cite{li2024can} leverages critic models to iteratively guide retrieval and reasoning toward task-specific goals. The critic model operates by evaluating potential sub-goals and their executions, assigning rewards to guide the selection of the most promising reasoning paths. This guidance ensures that the reasoning process aligns with task objectives, effectively navigating complex problem spaces~\cite{li2024can}. For domain-specific adaptation, SimRAG~\cite{xu2024simrag} employs self-training, generating and filtering synthetic data to fine-tune models for specialized fields. In biomedical applications, Self-Rewarding Tree Search (SeRTS)~\cite{hu-etal-2024-serts} combines Monte Carlo Tree Search and Reinforcement Learning to optimize retrieval. Speculative RAG~\cite{wang2024speculative} improves efficiency with a two-stage process: a smaller model drafts responses, while a larger model evaluates and finalizes them. This two-step process allows the system to balance efficiency and accuracy by leveraging the strengths of both models.

These approaches offer distinct benefits and limitations. SeaKR and Self-RAG provide dynamic adaptability and accuracy but demand significant computational resources. CR-Planner and SeRTS enhance task-specific precision but increase complexity. SimRAG excels in domain-specific tuning, however it is constrained by the need for high-quality synthetic data. Speculative RAG effectively reduces latency through parallel drafting and verification, but requires accurate evaluation by generalist models. 

{\em c) Long-Term Memory for Knowledge Retrieval:} Long-term memory integration in RAG frameworks addresses the limitations of purely query-specific retrieval by enabling the retention and reuse of knowledge across tasks. HippoRAG~\cite{gutierrez2024hipporag}, inspired by the hippocampal indexing theory, integrates long-term memory by linking a knowledge graph to an LLM and prioritizing relevant nodes using the Personalized PageRank algorithm. This approach allows the model to retrieve interconnected knowledge dynamically, consolidating past context for tasks like multi-hop reasoning, achieving up to 20\% better performance. It excels in repetitive and longitudinal tasks by enabling adaptive and context-aware retrieval, mimicking how the brain organizes and recalls episodic memories.

Various architectures embed long-term memory into RAG. MemLong~\cite{wang2024augmenting} employs a dual-network design where a frozen LLM backbone serves as a memory encoder, while a residual side-network manages retrieval, enabling efficient caching and updating of extensive contexts (up to 65k tokens). Its key advantage is scalability without data staleness, though managing large contexts may introduce overhead. HAT~\cite{aadhithya2024enhancing} introduces a Hierarchical Aggregate Tree structure that organizes dialogue history into a tree, where each node represents aggregated information from its child nodes. This design allows the system to manage extensive conversational contexts by traversing the tree to retrieve relevant information, enhancing coherence and summary quality. The recursive aggregation enables the model to handle long-term dependencies effectively, though challenges may arise in balancing tree depth with performance. MemoRAG~\cite{qian2024memorag} combines a lightweight global memory model with a retrieval-generation module, using draft answers as “clues” to guide precise retrieval from extensive datasets. It efficiently handles up to one million tokens by separating memory updates from retrieval operations, ensuring scalability and contextual relevance. This architecture excels in complex tasks but requires fine-tuning to balance memory efficiency and retrieval accuracy. Pistis-RAG~\cite{bai2024pistis} introduces a scalable, multi-stage framework for retrieval-augmented generation (RAG) systems, emphasizing alignment with human preferences through online learning and user feedback. Its architecture comprises distinct stages—matching, pre-ranking, ranking, reasoning, and aggregating—each refining the retrieval process to enhance response quality. A notable innovation is the ranking stage, which considers both semantic relevance and LLM preferences, addressing the sensitivity of LLMs to prompt ordering. By adopting a content-centric approach, Pistis-RAG integrates user feedback to continuously adapt and align with evolving user needs, resulting in a 9.3\% performance improvement on the MMLU (English) benchmark. However, the reliance on continuous user feedback may introduce variability, necessitating careful system tuning to maintain consistent performance.




{\em d) Multi-Hop and Multi-Modal Reasoning Retrieval:} Multi-hop and multi-modal reasoning broaden Retrieval-Augmented Generation (RAG)’s capacity to tackle tasks requiring complex, step-by-step deliberation and integration of diverse data sources. Multi-hop reasoning connects information across multiple steps to derive coherent answers, while multi-modal reasoning combines data from various formats such as text, images, and audio. This systematic approach enhances RAG’s ability to deliver comprehensive and well-founded responses to multifaceted queries.

Multi-layered Thoughts Enhanced RAG (METRAG)~\cite{gan2024similarity} integrates similarity- and utility-based reasoning for deeper contextual understanding. It does so by combining similarity-oriented retrieval with utility-oriented assessments, where a utility model, supervised by an LLM, evaluates the usefulness of retrieved documents beyond mere similarity using metrics like task relevance, informativeness, query-specific novelty, and completeness, enhancing the relevance and quality of the information utilized in generation.

RAG-Star~\cite{jiang2024rag} integrates retrieval augmentation with Monte Carlo Tree Search (MCTS) to improve problem-solving accuracy by iteratively planning intermediate sub-queries. Retrieval augmentation enables the model to incorporate external information, enhancing its reasoning process. Using MCTS, RAG-Star systematically explores reasoning paths by generating and evaluating intermediate sub-queries and their potential answers. This approach balances exploration and exploitation to identify the most promising reasoning trajectories, guiding the model toward highly accurate and contextually relevant solutions.

Knowledge Graph-Enhanced RAG Framework (KRAGEN)~\cite{matsumoto2024kragen} employs a “Graph-of-Thoughts” methodology to decompose multi-hop reasoning problems into explainable and systematic components. This approach structures the reasoning process by representing knowledge as interconnected concepts and relationships, often derived from knowledge graphs. During problem-solving, the model constructs this graph dynamically, allowing it to break down complex queries into smaller, manageable sub-tasks. By systematically addressing each component, KRAGEN enhances both the interpretability and accuracy of its reasoning process, providing a more transparent and effective framework for handling intricate queries.

However, a potential limitation of KRAGEN is its reliance on the quality and comprehensiveness of the underlying knowledge graph. If the knowledge graph lacks certain information or contains inaccuracies, the model’s reasoning and outputs may be adversely affected. Ensuring the knowledge graph is up-to-date and accurately reflects the domain is crucial for maintaining the effectiveness of the KRAGEN framework.


Building upon the foundational concepts of multi-hop and multi-modal reasoning, recent research has proposed innovative frameworks to address the inherent complexities of such tasks. These advancements focus on refining the step-by-step reasoning process and integrating diverse modalities, enabling models to navigate intricate queries with enhanced accuracy and explainability. By tackling challenges in sequential inferencing and combining textual with visual or other modal data, these frameworks set a new benchmark for retrieval-augmented generation systems

For instance, MultiHop-RAG provides a dedicated dataset and benchmarks to rigorously assess RAG systems on multi-step queries~\cite{tang2024multihop}, facilitating the evaluation of retrieval-augmented generation models in scenarios that necessitate reasoning across multiple documents. Retrieval-Augmented Multi-modal Chain-of-Thoughts Reasoning~\cite{liu2023retrieval} extends Chain-of-Thought (CoT) approaches to handle images and text in tandem, enabling models to process and reason over visual and textual data simultaneously. For purely textual multi-hop question answering, HOP, UNION, GENERATE (HUG)~\cite{zhao2023hop} offers a three-step method that models rationales as sets of sentences, enhancing explainability without requiring explicit rationale supervision. In this framework, "Hop" involves selecting relevant sentences, "Union" aggregates these sentences into a coherent rationale set, and "Generate" produces the final answer based on the aggregated rationale. The rationales modeled are the sets of sentences that collectively support the answer, providing transparency in the reasoning process by explicitly outlining the evidence considered. Multimodal-CoT and Multi-Chain Reasoning (MCR)~\cite{zhang2023multimodal} further advance reasoning by respectively separating rationale generation from answer inference for science question answering, and by prompting LLMs to examine multiple parallel chains of thought before synthesizing final solutions. These approaches address complex reasoning types that require integrating diverse information sources and evaluating multiple reasoning pathways. The rationale generated includes intermediate reasoning steps that elucidate the thought process leading to the answer. Prompting is generated by designing specific instructions that guide the model to consider various perspectives and reasoning chains, thereby enhancing the robustness and accuracy of the final output.

Although RAG improves factual correctness and can help the model explore a broader concept space by tapping into external repositories, it still does not imbue the model with a genuine, internal problem-solving strategy. 

{\em e) Self-Reflection Methods}: Recent advancements underscore the value of LLMs engaging in reflective reasoning before generating a final answer. Reflective reasoning involves the model's introspection and evaluation of its own thought processes to enhance decision-making and output quality.

Implicit Retrieval-Augmented Generation (RAG)~\cite{lewis2020retrieval,vatsal2024can,vatsal2024can2} instructs LLMs to first retrieve key chunks of context, specifying the number of sections and words in each section, then use these snippets to answer queries. The selection of the number of snippets and their lengths is typically determined through empirical tuning, balancing the need for comprehensive context with the constraints of the model's input capacity. This method has achieved near state-of-the-art results in both general and biomedical contextual question-answering tasks.

Metacognitive Prompting (MP)~\cite{wang2023metacognitive} draws on the concept of metacognition, comprising five phases:

1. \textit{Interpreting the Input}: The model analyzes the input text to grasp its context and meaning, ensuring a clear understanding of the task at hand. This is implemented by prompting the model to restate or summarize the input, confirming comprehension.

2. \textit{Forming an Initial Judgment}: Based on the interpreted input, the model generates a preliminary response or hypothesis, reflecting its immediate understanding. This involves producing an initial answer without external validation.

3. \textit{Critically Assessing that Judgment}: The model evaluates its preliminary response, identifying potential errors or uncertainties. This is achieved by prompting the model to question its initial answer, consider alternative interpretations, and assess the confidence level of its response.

4. \textit{Presenting a Final Decision with Reasoning}: After critical assessment, the model formulates a refined answer, providing a rationale that outlines the reasoning process. This step ensures transparency and allows users to understand the basis of the model's conclusion.

5. \textit{Gauging Confidence in the Entire Process}: The model reflects on the overall process, assigning a confidence score to its final answer, indicating the reliability of the response. This is implemented by having the model express its certainty level, guiding users in decision-making.

MP consistently outperforms Chain-of-Thought (CoT) and Plan-and-Solve methods across paraphrasing, natural language inference, and relation extraction tasks.


{\em f) Self-Critique Methods/Evaluation- and Verification-Focused Methods:} To enhance reliability and reduce factual inaccuracies in automated reasoning, self-critique methods have emerged as critical tools. These methods address challenges in producing consistent, accurate outputs by systematically verifying and refining initial responses.
Chain-of-Verification (CoVe)~\cite{dhuliawala2023chain} uses a four-step process: (1) generating an initial response, (2) formulating verification questions to identify potential errors or inconsistencies, (3) answering these questions to produce supporting evidence or rationale, and (4) revising the original response based on validated findings. CoVe has demonstrated over 10\% performance improvements compared to basic prompting and Chain-of-Thought (CoT) methods in both context-free and contextual question-answering tasks.

Verify-and-Edit (VE)~\cite{zhao2023verify} enhances uncertain CoT outputs by integrating external knowledge from reliable sources such as encyclopedias, knowledge graphs, or domain-specific repositories. Self-consistency identifies weak points in reasoning by generating multiple reasoning paths for the same problem and comparing their outputs for discrepancies or logical contradictions, revealing areas of low confidence or errors. The response is then revised by incorporating validated evidence, ensuring factual accuracy and logical coherence. Cross-referencing further verifies the revised response by re-checking it against retrieved knowledge to confirm it resolves inconsistencies while maintaining alignment across all reasoning steps, avoiding the introduction of new errors or contradictions. VE evaluates the reliability of the final output by analyzing agreement across revised reasoning paths and ensuring alignment with external knowledge. This approach has achieved up to 10\% gains in multi-hop reasoning tasks and 2\% improvements in truthfulness evaluations over CoT and self-consistency techniques.

In summary, self-critique methods, i.e., CoVe and VE, concentrate on verifying and refining initial outputs to reduce inaccuracies, while self-reflection techniques, e.g., Implicit RAG and MP, emphasize reflective reasoning for deepening understanding and clarity before producing an answer. CoVe and VE diverge in methodology: CoVe generates verification queries for self-checking, whereas VE specifically pinpoints uncertain outputs and edits them using external knowledge. 

\subsection{Reinforcement Learning} 

Reinforcement Learning (RL) provides a systematic framework for refining LLM behavior by guiding models toward desired objectives through iterative feedback and carefully designed reward signals from human feedback, automated metrics, or a pre-trained reward model~\cite{zhai2024fine}. There are six main components: agent, environment, state, action, reward, and policy~\cite{sutton2018reinforcement}. To apply RL for fine-tuning LLMs, the first step maps the six components to the LLM framework: the LLM represents the policy, while the current textual sequence is the state, and based on this state, the LLM generates an action, the next token. This action updates the state, creating a new state that incorporates the newly added token. After generating a complete textual sequence, a reward is determined by assessing the quality of the LLM output. This reward can be used to train a pre-trained reward model or can be directly integrated into the alignment process to guide the behavior of the model.

The RL methods adopted by these models can be divided into two main categories, model-based RL approaches and model-free approaches, which were discussed next.

{\em a) Model-based RL Approaches}: The methods in this category can be grouped into three categories, RLHF, RLAIF and exploration, which are discussed next.

Reinforcement Learning from Human Feedback (RLHF):  RL from Human Feedback (RLHF) re-train LLMs by incorporating a reward signal derived from human evaluations. RLHFs perform three fundamental stages: They initially perform supervised fine-tuning (SFT) using labeled datasets, followed by training a reward model (RM) based on human-evaluated outputs, and finally use this reward signal to inform the model’s policy fine-tuning using the Proximal Policy Optimization (PPO) algorithm~\cite{schulman2017proximal}.



\cite{ouyang2022training} created fine-tuned models like InstructGPT using human feedback to better adhere to user instructions. Similarly, \cite{shen2023loose} and \cite{wang2024secrets} explored reward modeling and methods to address challenges such as length bias, ensuring outputs are concise and aligned with human expectations. Frameworks like trlX \cite{havrilla2023trlx} and high-quality datasets introduced by \cite{cui2023ultrafeedback} have scaled RLHF applications, improving the performance of LLMs in tasks such as summarization, translation, and dialogue generation. Summarization tasks, for example, leverage reinforcement learning (RL) through both extractive and abstractive methods; extractive summarization selects key sentences from the source, while abstractive summarization generates novel sentences to convey the essence of the content \cite{stiennon2020learning}. RL optimizes summarization by using rewards based on metrics like ROUGE to iteratively enhance the quality of outputs. Policy optimization, on the other hand, employs pairwise feedback, comparing response pairs to align LLM outputs with human preferences. Techniques such as Pairwise Proximal Policy Optimization simplify the process by directly operating on comparative rewards, avoiding complexities like value function estimation and normalization \cite{munos2023nash}.

PPO algorithms iteratively adjust the weights of a model to maximize the expected reward~\cite{schulman2017proximal}. Central to this process is the collection of human feedback, which is critical in training reward models. Studies, such as Skywork-Reward~\cite{liu2024skywork} and TÜLU-V2-mix~\cite{ivison2023camels}, utilize human preferences by curating datasets of ranked examples, enabling models to align more effectively with human judgments. Additionally, ~\cite{li2023tool} introduces tool-augmented reward modeling, integrating external resources like calculators and search engines to refine alignment. Recent generative reward models use synthetic preferences, which are artificially created by sampling and ranking model outputs using a base preference model, to reduce reliance on extensive human feedback. ~\cite{scheid2024optimal} examined efficient methods for collecting pairwise human preferences, optimizing reward model design within RLHF frameworks. Additionally, research on over-optimization risks underscores the importance of balanced training to prevent performance degradation~\cite{gao2023scaling}. \cite{munos2023nash} propose novel pairwise feedback pipelines that improve preference learning and policy optimization by comparing response pairs to better capture human preferences.
    
RLHF’s multi-step process remains resource-intensive and reliant on extensive human feedback~\cite{kaufmann2023survey}. Over-optimization risks may cause models to exploit weaknesses in the reward function rather than achieving genuine alignment with human preferences~\cite{gao2023scaling}. 

RL from AI Feedback (RLAIF) is a training method designed to replace human evaluators with AI systems, offering better scalability and consistency by mitigating the variability of human judgment~\cite{lee2023rlaif}. In RLAIF, a Reward Model (RM) is trained using preference labels generated by an LLM. These labels are transformed into a probability distribution through a softmax function and optimized via cross-entropy loss, enabling the RM to guide the training of the target AI model~\cite{li2024hrlaif}. Various approaches have been proposed to address the specific challenges of RLAIF. For example, UltraFeedback compiles a large-scale dataset of over one million GPT-4 feedback annotations on 250,000 user-assistant conversations to train reward models~\cite{cui2023ultrafeedback}. Similarly, Magpie employs a self-synthesis method, where an aligned LLM generates large-scale alignment data that fine-tunes reward models~\cite{xu2024magpie}. HelpSteer2 introduces a permissively licensed preference dataset to train reward models, demonstrating improved alignment with human preferences~\cite{wang2024helpsteer2}. Another approach focuses on prompting LLMs to function as reward functions, directly guiding model training through reward scores, as seen in Exploring with LLMs (ELLM) Rewards~\cite{du2023guiding}. Additional work, such as Reward Design with Language Models, emphasizes constructing reward mechanisms that align model outputs with desired outcomes by leveraging LLM capabilities~\cite{kwon2023reward}. Self-supervised feedback mechanisms have also been explored; for instance, the Eureka framework introduces a novel approach to reward optimization through self-generated feedback loops~\cite{ma2023eureka}. Self-rewarding systems, including Self-Refined LLMs ~\cite{song2023self} and Self-Rewarding Language Models (SRLM) ~\cite{yuan2024self}, enable iterative refinement of model outputs based on their own evaluations.

Despite its potential, RLAIF remains less widely adopted compared to RLHF. This discrepancy stems from challenges, such as difficulties in achieving alignment and the risk of propagating biases inherent in AI-generated feedback~\cite{cui2023ultrafeedback, ma2023eureka}. These challenges can create feedback loops that amplify existing biases, constraining model diversity and limiting its ability to generalize effectively~\cite{yuan2024self}. Moreover, the absence of human evaluators in RLAIF can result in a lack of nuance, leading to a narrower latent space influenced by the biases of the training AI~\cite{song2023self}. 

Exploration techniques in RL involves seeking new information to improve future decisions, whereas exploitation capitalizes on current knowledge to maximize immediate rewards~\cite{wang2018exploration}. In these algorithms, each action decision can be made stochastic via epsilon-greedy ~\cite{dann2022guarantees} or entropy regularization ~\cite{mnih2016asynchronous} to ensure diverse coverage of the environment, but excessive exploration can be inefficient. Traditional approaches, such as epsilon-greedy ~\cite{tokic2010adaptive} and Boltzmann exploration ~\cite{cesa2017boltzmann}, introduce randomness without leveraging prior knowledge, slowing convergence. Recent methods, like, ExploRLLM~\cite{ma2024explorllm} presents a hierarchical reinforcement learning framework that combines the strengths of LLMs and affordance-based policies. In this approach, LLMs generate high-level plans to outline strategic goals, while affordance-based policies, which identify actionable possibilities within the environment, execute specific actions to achieve those goals. This method improves exploration efficiency by prioritizing high-value states and reducing reliance on frequent LLM invocations. Despite its effectiveness in structured environments, the approach faces challenges in adapting to dynamic and open-ended domains~\cite{zhao2024epo}. 
    

Soft RLLF~\cite{nguyen2024balancing} integrates natural language as logical feedback to balance exploration and exploitation, enabling improved performance in reasoning tasks such as negation understanding and logical consistency in high-stakes applications. This is achieved by encoding logical consistency checks and negation handling into the learning process, utilizing feedback loops to iteratively refine the agent’s decision-making. However, its effectiveness diminishes when tackling problems requiring broader adaptability and creativity, as it is optimized for structured reasoning~\cite{nguyen2024balancing}. Another recent approach, LLM+Exp~\cite{yang2023empower}, employs a dual-LLM framework: one LLM analyzes action-reward trajectories to derive exploration strategies, while the other dynamically adjusts action probabilities to refine future decisions. Action-reward trajectories represent sequences of actions taken by an agent and the corresponding rewards, offering insights into the learning process. Action probabilities define the likelihood of selecting specific actions based on learned patterns and anticipated outcomes. While this adaptive approach excels in structured environments, it faces scalability issues and struggles to generalize effectively to unpredictable or unstructured tasks. 
Guided Pretraining RL~\cite{du2023guiding} leverages LLMs to enhance exploration by providing contextual background knowledge. This knowledge helps prioritize relevant actions, improving sample efficiency—allowing the agent to learn more effectively from fewer interactions. The method involves generating structured trajectories, which represent meaningful sequences of actions related to the task, using LLMs. These trajectories are used to pretrain the agent, providing a solid foundation before fine-tuning its policies in the target environment.

While this approach excels in providing structure and reducing exploration costs, it struggles with tasks that require broader adaptability and creative problem-solving. The reliance on predefined trajectories limits its ability to generalize to highly variable or unpredictable environments, where more flexible reasoning is needed.

{\em b) Model Free Approaches}: These methods can be grouped into three categories, DPO, IPO, and actor critical. Their discussion follows next.
    
Direct Preference Optimization (DPO) addresses the limitations of RLHF/PPO, which necessitates meticulous oversight and significant computational resources due to the initial phase to train a reward model using a preference dataset, followed by training an RL policy with the pre-trained reward model serving as the environment. DPO offers a simpler alternative by directly optimizing LLM parameters using preference data, bypassing the need for a reward model~\cite{rafailov2024direct}. DPO relies on a preference loss function trained on datasets of paired human preferences (e.g., "Response A is better than Response B"). Several extensions to DPO improve upon this baseline. For instance, DPOP \cite{pal2024smaug} (also termed DPO-positive) introduces a margin-based term to prevent rewarding both preferred and disfavored outputs concurrently, thereby improving performance on tasks with small edit distances. Specifically, the margin-based term in DPOP introduces a penalty for assigning high probabilities to both preferred and disfavored outputs, ensuring that the model distinctly favors the preferred response to improve task performance. Iterative DPO~\cite{yuan2024self} (also known as \emph{online} DPO) mitigates distribution shifts by continually updating the policy on newly generated responses, an advantage over vanilla DPO, which can overfit to a narrower distribution. Meanwhile, $\beta$-DPO~\cite{wu2024beta} adaptively tunes the regularization term based on the data quality, making it more robust to noisy preferences. Stepwise DPO (sDPO)~\cite{kim2024sdpo} partitions the preference dataset to perform incremental updates, leveraging a stronger intermediate reference model at each phase.

DPO methods are particularly advantageous for structured problem-solving, like in creative writing or complex reasoning because they can directly incorporate human preferences and avoid undesired behavior without heavily relying on large-scale reward modeling or complex RL training loops~\cite{rafailov2024direct}. However, a recurring drawback is their sensitivity to distribution shifts, e.g., when the model starts generating out-of-domain responses, alignment performance can drop unless the reference model or preference data is iteratively updated~\cite{yang2024weighted}. Moreover, purely relying on pairwise or setwise human judgments can still introduce label noise or ambiguity, especially for creative or unstructured tasks~\cite{chowdhury2024provably}. Despite these limitations, DPO-based techniques are promising for balancing helpfulness and correctness in open-ended LLM outputs
~\cite{xiao2024comprehensive}.

Identity Preference Optimization (IPO) \cite{azar2024general} was introduced to address the overfitting inherent in RLHF and DPO. Unlike traditional methods that transform pairwise preferences into pointwise rewards using the Bradley–Terry (BT) model~\cite{sun2024rethinking}, IPO directly optimizes preferences without relying on nonlinear transformations, which are known to exacerbate overfitting. The objective function of Identity Preference Optimization (IPO), as defined in eq.~(\ref{eq:IPO_general}), aims to directly optimize preference probabilities while mitigating overfitting issues inherent in methods like RLHF and DPO. The function maximizes the expected preference utility, represented by 
\[
\mathbb{E}_x \left[ \mathbb{E}_{y, y'} \Psi(P_\theta(y > y'))\right],
\]
where \(\Psi(P_\theta(y > y'))\) captures the model's ability to predict and optimize preference probabilities for pairs of outputs (\(y\) and \(y'\)). To prevent excessive deviation from a reference policy, the KL divergence term \(D_{\text{KL}}(\pi || \pi_\text{ref})\) imposes a regularization constraint, controlled by the coefficient \(\beta\). By balancing preference optimization and regularization, this approach avoids transforming pairwise preferences into pointwise rewards, which can exacerbate overfitting, and directly aligns the model’s behavior with human preferences while maintaining stability. 
\begin{equation}
\pi^*_\theta = \max_\pi \mathbb{E}_{x} \left[ \mathbb{E}_{y, y'} \Psi(P_\theta(y > y')) - \beta D_{\text{KL}}(\pi || \pi_\text{ref}) \right].
\label{eq:IPO_general}
\end{equation}



To address the overfitting caused by the nonlinear transformation \(\Psi(x)\), IPO simplifies \(\Psi(x)\) to a linear function, \(\Psi(x) = x\), and formulates a robust loss function, as defined in eq.~(\ref{eq:IPO_loss}). This loss function, \(L_{\text{IPO}}\), directly optimizes the policy \(\pi_\theta\) by aligning it with human preferences while mitigating overfitting. The expectation is taken over pairs of outputs \((y_w, y_l)\), where \(y_w\) represents the preferred (winning) output and \(y_l\) the less preferred (losing) output. The terms \(-\log \frac{\pi_\theta(y_w)}{\pi_\text{ref}(y_w)}\) and \(-\log \frac{\pi_\theta(y_l)}{\pi_\text{ref}(y_l)}\) measure how well the current policy \(\pi_\theta\) aligns with the reference policy \(\pi_\text{ref}\), accounting for both preferred and less preferred outputs. A regularization term, \(\frac{1}{2\beta}\), balances the trade-off between optimizing preferences and maintaining adherence to the reference policy, ensuring model stability and reducing the risk of overfitting. By incorporating a squared penalty term, \(L_{\text{IPO}}\) captures and penalizes deviations from ideal preference alignment, whether positive or negative. The simplified approach avoids the complexity and instability of nonlinear transformations, providing a stable and effective framework for aligning policies with human preferences. This makes IPO a robust and efficient alternative to traditional preference-based learning methods that rely on pointwise rewards or complex transformations.
\begin{equation}
L_{\text{IPO}} = - \mathbb{E}_{(y_w, y_l)} \left[ \log \frac{\pi_\theta(y_w)}{\pi_\text{ref}(y_w)} - \log \frac{\pi_\theta(y_l)}{\pi_\text{ref}(y_l)} - \frac{1}{2\beta} \right]^2.
\label{eq:IPO_loss}
\end{equation}

This approach proves particularly robust in scenarios with deterministic or near-deterministic feedback, where existing methods often struggle due to unstable gradients~\cite{wu2024self}. By leveraging a simpler optimization framework and incorporating strong regularization, IPO effectively mitigates overfitting and outperforms DPO in experimental settings~\cite{wang2024preference}. However, IPO faces challenges due to its reliance on static preference distributions, which limits adaptability to dynamic or diverse scenarios. Additionally, its sensitivity to noise and dependence on high-quality data reduce robustness in complex, evolving environments~\cite{fisch2024robust}.


Actor-critic methods, such as Advantage Actor-Critic (A2C) and Deep Deterministic Policy Gradient (DDPG), have been effectively adapted to optimize prompts for LLMs. Frameworks like Prompt Actor-Critic Editing (PACE)~\cite{dong2023pace} employ an iterative process where the \textit{actor} (the LLM) generates a response \(a\) based on a prompt \(p\) and input \(X\). This process is formalized as 
\[
a = f_{\text{actor}}([p; X], M),
\]
where \(f_{\text{actor}}\) represents the decision-making mechanism of the actor, \([p; X]\) is the concatenated context consisting of the prompt \(p\) and the specific input \(X\), and \(M\) is the LLM being optimized. The actor function processes the concatenated context to produce the response \(a\), guided by the prompt \(p\) and the input \(X\).

The critic, another LLM or evaluation mechanism, evaluates the relevance, coherence, and task-specific accuracy of the response against the objective \(Y\). The critique is calculated as follows~\cite{dong2023pace}:
\[
c = f_{\text{critic}}([p; X; a; Y], M),
\]
where \(f_{\text{critic}}\) represents the evaluation function of the critic. The input \([p; X; a; Y]\) consists of the prompt \(p\), the input \(X\), the actor-generated response \(a\), and the objective \(Y\), which defines the desired or target output. The critic processes this concatenated input using the language model \(M\) to generate a critique \(c\). This critique assesses how well the response \(a\) aligns with the objective \(Y\), considering both the input \(X\) and prompt \(p\). \cite{ziegler2019fine} leverages KL-regularization to balance fidelity to the original prompt while allowing modifications that improve task-specific performance. By iterating on this actor-critic loop, PACE enhances prompt effectiveness and guides LLMs toward better alignment with task objectives. 

Additionally, actor-critic methods assume well-structured feedback loops, which might be unreasonable for problems with sparse or noisy signals. Recent work addresses these challenges. \cite{meier2022open} explores open-ended learning in the context of unsupervised skill discovery, highlighting the need for more flexible reward functions in high-dimensional environments. HDFlow~\cite{yao2024hdflow} combines fast and slow thinking modes to enhance complex reasoning. \cite{liu2024enhancing} introduces Direct Q-function Optimization (DQO), which formulates response generation as a Markov Decision Process (MDP), allowing each token generation to be treated as a state transition. Leveraging the soft actor-critic (SAC) framework, DQO directly parameterizes the Q-function within the language model, enabling it to learn effectively from offline data, including unbalanced or negative samples that helps improve multi-step reasoning. 

\section {Similarities of LLMs and Traditional Automated Implementation Generation Methods and Related Research Needs}


A broad analogy can be identified between using Genetic Algorithms (GAs) and LLMs for implementation creation. 
\begin{enumerate}
\item 
\textbf{Selection:} GA selection chooses the fittest individuals to pass their genes to the next generation. In fine-tuning or training data selection, LLMs prioritize coherence and relevance when generating textS, similar to selecting relevant context for responses. Like choosing the best seeds from a harvest, LLMs select the most relevant words or sentences to continue a conversation~\cite{aryan2024llms}.

\item 
\textbf{Crossover (Recombination)}: GA crossover combines the genomes of two parents to create a new individual. This is similar to blending knowledge from different domains during text generation. For example, merging insights from literature and science in a single response. Crossover is like an LLM writing poetry about quantum physics, e.g., combining Shakespearean elegance with scientific rigor~\cite{wu2024evolutionary}.

\item 
\textbf{Mutation:} GA mutation introduces random changes in a genome to explore new possibilities. Similar to the slight randomness added during sampling techniques like top-k or temperature settings, which allow LLMs to produce diverse responses. Mutation in GAs is like LLMs occasionally breaking patterns to say something unexpected or creative~\cite{yin2024controlling}.

\item 
\textbf{Inversion:} GA inversion reverses a segment of the genome to explore new configurations. This parallels rephrasing or reordering sentences during text generation while preserving the original meaning. Like flipping a playlist order for a new vibe, LLMs rephrase “The car is fast” into ``A fast car it is''~\cite{guo2023connecting}.

\item \textbf{Elitism:} GA ensures the best solutions carry over unchanged to the next generation. Similar to checkpointing the best-performing weights during training or favoring high-confidence outputs in decoding strategies. Like archiving the best answers during an essay edit, LLMs retain their most confident responses for the final output~\cite{brahmachary2024large}.

\item 
\textbf{Replacement:} GA decides how much of the old population to keep versus the new one. Similar to parameter updates during fine-tuning, where new knowledge replaces older information incrementally. Replacement is like LLMs balancing old facts while integrating new updates, ensuring a model doesn’t ``forget'' but adapts to current knowledge~\cite{chao2024match}.

\item 
\textbf{Fitness Evaluation:} GA scores individuals based on quality to determine their survival. Similar to evaluating model outputs using metrics like BLEU, ROUGE, or user feedback in RLHF. Fitness evaluation is like an LLM receiving human feedback to improve its responses based on relevance, coherence, or creativity \cite{hao2024large}.

\item 
\textbf{Exploration vs. Exploitation:} GA balances trying new possibilities (exploration) and refining known solutions (exploitation). Balancing randomness and coherence during response generation. Parameters like temperature encourage exploration, while context relevance drives exploitation. Just as genetic algorithms search for novel solutions, LLMs strike a balance between playful creativity and logical reasoning in ambiguous prompts \cite{huang2024wese}.
\end{enumerate}

\begin{figure}[!t]
\includegraphics[width=0.46 \textwidth]{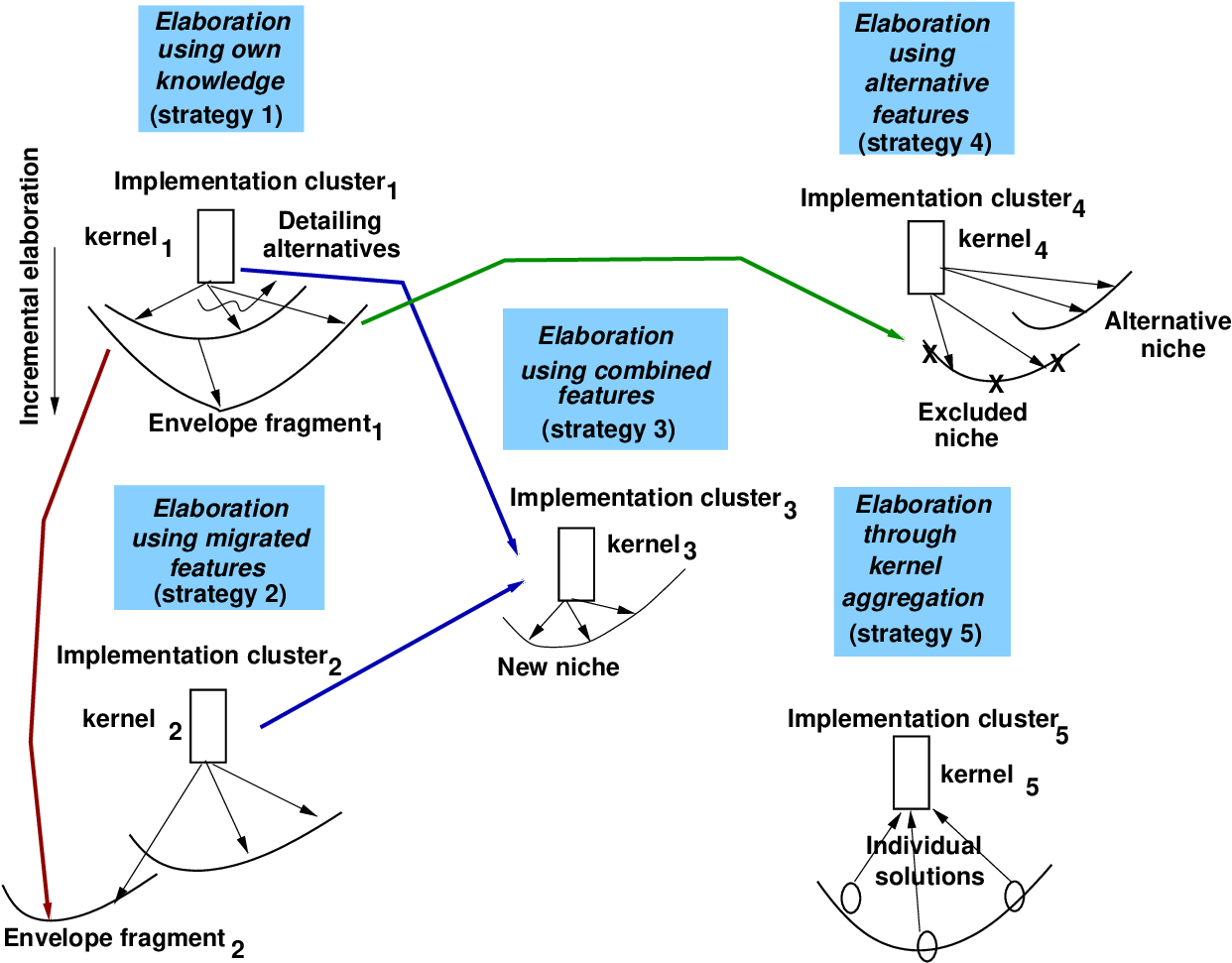}
\caption{Five strategies for automated implementation creation}
\label{fig1}
\end{figure}

The next part discusses using Cognitive Architectures for implementation creation, possibly using the features of LLMs. Similar to~\cite{Li2018}, this report considers that devising an implementation for a problem specification utilizes the five strategies shown in Figure~\ref{fig1}. The problem solving process is a mixture of the five strategies. 

Each strategy starts from a {\em kernel}, which is the invariant set of features used in the process. The problem solving process creates a solution cluster corresponding to the kernel features, e.g., each implementation in the cluster includes the features. Implementations are created through implementation elaboration by exploring a sequence of detailing alternatives. For example, the principle of the bubble sorting algorithm can be described as repeatedly comparing the adjacent values of an array and swapping them if they are in the wrong order until no more value swapping are needed. The kernel features include three features: (i)~the values of an array, (ii)~the swapping of adjacent values if they are in the wrong order, and (iii)~the repetition of the process until no more swapping are needed. The corresponding cluster includes all implementations obtained by elaborating the three kernel features.  

The five strategies are as follows~\cite{Li2018}:
\begin{itemize}
\item 
Strategy 1 describes the elaboration process in which each kernel is elaborated without changing the kernel. A set of detailing alternatives can be used for each elaboration step to produce an implementation envelope. The envelopes are incrementally elaborated until the final implementation is created.    

\item 
Strategy 2 represents the process, which in addition to the elaboration steps of Strategy~1 also uses elaboration results corresponding to a different implementation cluster. Figure~\ref{fig1} shows the using of features from Implementation cluster~1 (red arrow in the figure) to build the implementations of Implementation cluster~2. Hence, the subsequent solution include elaboration of all kernel features and the features adopted from another cluster. 

\item 
Strategy 3 uses a kernel that combines kernel features from two different implementation clusters. The blue arrows in Figure~\ref{fig1} illustrates the combination.  

\item 
Strategy 4 presents an elaboration process in which the selected detailing alternatives are excluded from the elaboration steps used for building other implementation clusters. It represents the excluded niche in Figure~\ref{fig1} (green arrow).  

\item
Strategy 5 creates a kernel bottom-up by identifying and generalizing the features of individual implementations. The individual implementations were produced through less-structured methods, like, for example, through experimental trial-and-error.  \end{itemize}

While the five strategies provide templates for the implementation elaboration process, automated implementation generation requires the following additional activities:
\begin{enumerate}
\item 
{\em Divide and conquer}: The activity partitions a problem into sub-problems and then provides ways to integrate the implementation for the sub-problems. Task decomposition methods in LLM prompting~\cite{zhou2022least, khot2022decomposed} can produce certain decompositions, especially in situations for which the sb-problems are less coupled. However, design problems are often strongly coupled, so that even though there are specialized modules to implement a certain function, their operation and performance are tightly related. Decomposition requires not only a static problem partitioning based on the items in the prompts (i.e. words) but also the interpretation of a sub-problem within the context set-up by the interpretations of other sub-problems. LLM fine tuning through RL is likely infeasible due to the huge space of possible decompositions possible in real-life. A mechanism is also needed to track the analyzed decompositions, so that the information can be used to improve future decompositions. This capability is absent in current methods.  

\item
{\em Kernel creation}: The method creates kernels either by assembling the features likely to address the problem requirements and then elaborating them top-down or in a bottom-up process as detailed in Strategy 5. Separate kernels can be created for different sub-problems followed by integrating them into a single kernel and its elaboration or separately elaborating each kernel and integrating their implementations. Ideas on LLM self-reflection and focus on the main information~\cite{weston2023system} can help identify the features to be included in a kernel. However, finding kernels, e.g., the invariant features present in all implementations pertaining to a cluster, remains mostly a manual process. Methods similar to RLHF~\cite{schulman2017proximal} can help retrieving similar features, but their scalability is likely low. Moreover, combining features from different kernels to generate a new kernel (Strategy~3) has not been studied by current LLM methods. The combination of features needs a way to predict the expected performance at a high level (possibly a qualitative evaluation), which can be offered to some degree by LLM, similar to the use of LLM to solve ambiguities~\cite{liu2024gram, chen2024tablerag}. However, it is likely that the current methods are insufficient for this purpose. 

\item
{\em Elaboration}: Executing the five strategies requires devising additional methods for detailing alternatives, predicting the effectiveness of each alternative in the context of a partial implementation, assigning a priority (preference) to each alternative, and incorporating the alternative into the partial implementation. A possible approach is to use schema for elaboration, similar to RAG methods for LLMs~\cite{dixit2024sbi, matsumoto2024kragen}. Schema matching can benefit from LLMs to clarify certain ambiguities, such as in~\cite{liu2024gram, chen2024tablerag}. However, schema are static structures, useful in analogical reasoning, even though problem solving often requires performing new sequences of decisions beyond a static schema. 

\item 
{\em Implementation assessment}: LLMs can be used for two kinds of performance assessments. Qualitative assessment, including comparing implementations, such as pairs of circuit designs, can be obtained by prompting traditional LLMs. CoT prompting can be used to obtain performance assessment at a finer granularity. RLHF can fine tune assessment by adding human feedback about the quality of the implementations~\cite{cui2023ultrafeedback}. Moreover, self-critique methods could be used to improve the correctness and completeness of the LLM responses, like self-consistency and cross-referencing methods in VE~\cite{zhao2023verify}. A second approach uses datasets of characterized implementations to train an LLM, similar to exploration techniques in RL~\cite{tokic2010adaptive, cesa2017boltzmann}. Then, the generalization capacity of LLMs are used to quantitatively predict the performance parameters of a new implementation. Nevertheless, the two approaches do not scale beyond the samples used in training an LLM, including situations in which a new implementation uses a nonlinear combination of the features of different implementation. There is no mechanism similar to setting-up precise physical models of an implementation, so that the models can be solved to produce quantitative performance assessment, like in traditional automated implementation creation methods.      

\item 
{\em Memory and learning}: Similar to using long-term memory for knowledge retrieval in RAG, memory systems are needed to for learning to store associations, like kernel features, their most relevant implementations fragments, and their performance values or between high-level features and their detailed elaborations, the causal relationships of main features and performance attributes, and elaboration sequences that produced high-quality implementations. Similar to schema-based retrieval, memory cueing must solve semantic ambiguities.    

\item
{\em Adaptive process}: It includes the sequence of automated activities performed to create an implementation. It requires devising new means to predict the expected outcomes of the available activities, selecting and adapting an activity to the current context, understanding the degree to which the sequence advances towards creating an implementation, and learning new knowledge available during the process. Also, when addressing collaboration between humans and LLMs to tackle unexpected challenges, such as handling zero-day attacks, the process necessitates reasoning, understanding of prior instructions, and intuitive decision-making within the context of new parameters and constraints. To automate this process, exploring reasoning techniques, including deductive reasoning, inductive reasoning, analogical reasoning, common sense reasoning, tree-of-thoughts, multiple chains of thought, causal reasoning, heuristic reasoning, and symbolic reasoning, is required. Among these, the primary human thought process often involves mapping the current problem to a previously encountered one or identifying similarities with analogous problems, like in analogical reasoning. Consequently, an effective approach to problem modeling could involve neuro-symbolic representations that allow LLMs to dynamically learn and adapt in real-time. Techniques such as grokking~\cite{liu2022towards}, which enable models to discover relationships and patterns through iterative refinement, and masked LLMs are promising methods to achieve this goal. These approaches empower the model to derive connections on the fly, effectively merging learned representations with reasoning capabilities.

\end{enumerate}





\section {Conclusions}

Recent advances in Large Language Models (LLMs) offer the opportunity to extend automated implementation generation techniques beyond the current methods that require algorithmic specifications as input and can use only statically domain knowledge. LLMs can process multi-modal descriptions, including ideas communicated in natural language and using images and with certain degrees of specification completeness, unknowns, and ambiguity. LLMs learn a broad range of associations and for diverse contexts.  These new capabilities might offer intriguing paths beyond traditional implementation generation, such as support problem framing and exploration of possible solution approaches, 
improved implementation assessment across abstraction levels by comprehensive comparison to similar, externally available implementations, collective feedback and preferences, and enhanced elaboration by incorporating continuously updated domain knowledge. These features are critical in solving open-ended problem, currently hard to address with existing methods. Summarizing the state-of-the-art on LLMs and their related improvements is a first step towards devising nocel LLM-based methods for implementation generation. 

This report offers a comprehensive overview of existing LLM techniques and studied the degree to which they can model the activities needed for implementation generation for open-ended problem solving. The overview presents LLM enhancements, like
prompting, Reinforcement Learning (RL) and Retrieval-Augmented Generation (RAG). Then the report discusses the possibility of using LLMs to realize problem solving activities that are not available in traditional automated implementation generation methods. New research requirements are also presented, e.g., support for problem framing, creating an implementation approach, effective elaboration control, robust qualitative and quantitative assessment across abstraction levels, knowledge memorizing during learning, and managing the problem solving process.

\bibliography{citations}
\end{document}